\documentclass{article}
\usepackage{spconf,amsmath,epsfig}
\usepackage{amsthm}
\usepackage{amssymb}
\usepackage{bbm} 

\theoremstyle{my_style}
\newtheorem{theorem}{Theorem}
\newtheorem{lemma}{Lemma}
\newtheorem{proposition}{Proposition}

\newcommand{\cN}{{\mathcal{N}}}

\newcommand{\cH}{\mathcal{H}}

\newcommand{\cL}{\mathcal{L}}

\newcommand{\cD}{\mathcal{D}}

\newcommand{\bI}{\boldsymbol{I}}

\newcommand{\bx}{\boldsymbol{x}}

\newcommand{\by}{\boldsymbol{y}}
\newcommand{\bu}{\boldsymbol{u}}
\newcommand{\bv}{\boldsymbol{v}}
\newcommand{\bz}{\boldsymbol{z}}
\newcommand{\ba}{\boldsymbol{a}}

\newcommand{\bc}{\boldsymbol{c}}

\newcommand{\bw}{\boldsymbol{w}}

\newcommand{\bZero}{\boldsymbol{0}}

\newcommand{\bzero}{\boldsymbol{0}}

\newcommand{\bdelta}{\boldsymbol{\delta}}
\newcommand{\btheta}{\boldsymbol{\theta}}
\newcommand{\bomega}{\boldsymbol{\omega}}

\newcommand{\beq}{\begin{equation}}
\newcommand{\eeq}{\end{equation}}
\newcommand{\beqn}{\begin{eqnarray}}
\newcommand{\eeqn}{\end{eqnarray}}
\newcommand{\beqns}{\begin{eqnarray*}}
\newcommand{\eeqns}{\end{eqnarray*}}

\newcommand{\R}{\mathbb{R}}

\makeatletter
\newcommand{\bdiv}{\mathop{\operator@font div}}
\makeatother

\makeatletter
\newcommand{\diag}{\mathop{\operator@font diag}}
\makeatother

\makeatletter
\newcommand{\conv}{\mathop{\operator@font conv}}
\makeatother

\makeatletter
\newcommand{\minim}{\mathop{\operator@font minimize}}
\makeatother

\makeatletter
\newcommand{\maxim}{\mathop{\operator@font maximize}}
\makeatother

\makeatletter
\newcommand{\sign}{\mathop{\operator@font sign}}
\makeatother \makeatletter

\makeatletter
\newcommand{\proj}{\mathop{\operator@font proj}}
\makeatother \makeatletter

\makeatletter
\newcommand{\spa}{\mathop{\operator@font span}}
\makeatother \makeatletter

\makeatletter
\newcommand{\epi}{\mathop{\operator@font epi}}
\makeatother \makeatletter

\makeatletter
\newcommand{\dom}{\mathop{\operator@font dom}}
\makeatother \makeatletter

\makeatletter
\newcommand{\real}{\mathop{\operator@font Re}}
\makeatother \makeatletter

\makeatletter
\newcommand{\imag}{\mathop{\operator@font Im}}
\makeatother \makeatletter

\makeatletter
\newcommand{\sinc}{\mathop{\operator@font sinc}}
\makeatother \makeatletter

\makeatletter
\newcommand{\trace}{\mathop{\operator@font tr}}
\makeatother \makeatletter

\title{Efficient KLMS and KRLS Algorithms: A Random Fourier Feature Perspective}

\name{Pantelis Bouboulis, Spyridon Pougkakiotis, S. Theodoridis \thanks{This research was funded by the European Union (European
Social Fund - ESF) through the EC - FP7 FET program HANDiCAMS.}}
\address{University of Athens\\
    Department of Informatics and Telecomunications\\
    Athens, Greece.\\
    panbouboulis@gmail.com,  sdi1200151@di.uoa.gr, stheodor@di.uoa.gr}

\begin{document}
\ninept
\maketitle

\begin{abstract}
We present a new framework for online Least Squares algorithms for nonlinear modeling in RKH spaces (RKHS). Instead of implicitly mapping the data to a RKHS (e.g., kernel trick), we map the data to a finite dimensional Euclidean space, using random features of the kernel's Fourier transform. The advantage is that, the inner product of the mapped data approximates the kernel function. The resulting ``linear'' algorithm does not require any form of sparsification, since, in contrast to all existing algorithms,  the solution's size remains fixed and does not increase with the iteration steps. As a result, the obtained algorithms are computationally significantly more efficient compared to previously derived variants, while, at the same time, they converge at similar speeds and to similar error floors.
\end{abstract}

\section{Introduction}
\label{sec:intro}
Online learning in RKH spaces has attracted a lot of interest over the last years, see, e.g., \cite{Kivinen_2004_11230, Engel_2004_11231, Theo_ML, EREF2, Slavakis_2008_9257, Slavakis_2009_10648, Slavakis_2011_11460, Vaerenberg_KRLS}.
The Kernel Least Mean Square (KLMS) algorithm, introduced in \cite{Liu_2008_10645, Bouboulis_2011_10643}, presents a simple and efficient method to address non linear adaptive filtering tasks. Considering a sequentially arriving data of the form $\{(\bx_n, y_n),\; n=1,2,\dots\}$, where $\bx_n\in\R^d$, $y_n\in\R$, generated by a non-linear model, KLMS's mechanism can be summarized as follows: (a) map each arriving input datum, $\bx_n$, to an infinite dimensional Hilbert space $\cH$, using a specific kernel $\kappa$ and (b) apply the LMS rationale
to the transformed data, i.e., $\{(\kappa(\bx_n,\cdot), y_n),\; n=1,2,\dots\}$. Its main drawback is that the solution is given in terms of a linear expansion of kernel functions (centered at the input data points $\bx_n$), which grows infinitely large (proportionally to $n$), rendering its application prohibitive both in terms of memory and computational resources. The centers, $\bx_n$, that make up the linear expansion of the solution, are said to constitute the \textit{dictionary}.  In practice, sparsification methods are applied to keep the size of the dictionary sufficiently small and make the algorithm computationally tractable. These methods adopt a suitably selected criterion to decide whether a particular datum (i.e., $\bx_n$) will be included in the dictionary or not. Popular variations include the quantization \cite{QKLMS}, the novelty \cite{Liu_2008_10645}, the coherence \cite{RichBerm} and the surprise \cite{Liu_2010_10644} criteria.

Although the aforementioned sparsification techniques are able to reduce the size of the expansion significantly, they, too, require significant computational resources, even when the dictionary is small. This is due to the fact that at each iteration step, $n$, a sequential search over all the current dictionary elements has to be performed, in order to determine whether the new center, $\bx_n$, will be added to the dictionary or not.
Another important issue is the dimension of the input space. If this is small (e.g., $d<5$), then the aforementioned sparsification strategies may result in dictionaries with a few dozens elements, without compromising Mean Square Error (MSE) performance. However, if this dimension grows larger, then these methods will inevitably give dictionaries with several thousands elements or more rendering KLMS prohibitively demanding due to the sequential search over large dictionaries.  Furthermore, from a theoretical point of view, such approaches are not elegant, in the sense that they build around ``ad hoc" arguments, which, also, complicate the corresponding theoretical analysis.

The aforementioned difficulties have limited the extension of KLMS to more general settings, such as in distributed learning. In this case, the exchange of dictionaries among the network's nodes increase the network's load significantly \cite{Distributed_KLMS1, Distributed_KLMS2, Distributed_KLMS3}. More importantly, as each node should match its dictionary with the dictionaries of its neighbors (applying multiple sequential searches) the required computational resources  become quite demanding. In the present work, we follow a different rationale. Instead of mapping the input data to an infinite dimensional Reproducing Kernel Hilbert Space, induced by the selected kernel, and subsequently sparsifying the solution, we map the input data to a finite (although larger  than the input one) dimensional Euclidian space $\R^D$. However, this mapping is done in a sensible way that cares for a good {\it approximation} of the kernel evaluations. The mapping to $\R^D$ is carried out using random features of the kernel's Fourier transform \cite{RahimiRecht, DependentFourFeat, Sparse_GP}. Following this approach, the resulting algorithm, which we call \textit{Random Fourier Features Kernel LMS} or RFFKLMS for short, leads naturally to a standard linear LMS, with a \textit{fixed-size} solution   (i.e., a vector in $\R^D$); thus, no special sparsification techniques are needed.  RFFKLMS is computationally lighter than various variants of KLMS, while at the same time it exhibits the same MSE performance (for sufficient large $D$).  Similar arguments as before hold true for the case of the KRLS.

Section \ref{SEC:KLMS} briefly describes the rationale behind the standard KLMS with the quantization sparsification strategy. Sections \ref{SEC:RFF} and \ref{SEC:RFFKLMS} present the theory of approximating shift-invariant kernels with random features of their Fourier Transform and the new linearized implementation of the KLMS using this approximation. Simulations are given in section \ref{SEC:experiments}. Section \ref{SEC:krls} briefly describes the ``linearized'' version of KRLS based on the random Fourier features approximation framework, while section \ref{SEC:concl} concludes the paper. In the following, matrices appear with capital letters and vectors with small bold letters.

\section{The Quantized KLMS}\label{SEC:KLMS}
Consider the sequence $\cD = \{(\bx_n, y_n),\; n=1,2,\dots\}$, where $\bx_n\in\R^d$ and $y_n\in\R$. The goal of the KLMS is to learn a non-linear input-output map $f$, so that to minimize the MSE, i.e., $\cL(f) = E[(y_n - f(\bx_n))^2]$. Typically, we assume that $f$ lies in a RKHS induced by the Gaussian kernel, i.e., $\kappa_\sigma(\bu,\bv) = e^{-\|\bu-\bv\|^2_2/(2\sigma^2)}$, for some $\sigma>0$. Computing the gradient of $\cL$ and estimating it by its current measurement (as it is typically the case in LMS), we take the solution at the next iteration, i.e.,
$f_n = f_{n-1} + \mu e_n \kappa(\bx_n,\cdot)$,
where $e_n=y_n - f_n(\bx_n)$ and $\mu$ is the step-size (see \cite{EREF2, Liu_2010_10644} for more). Assuming that the initial solution is zero, the solution after $n$ steps becomes $f = \sum_{i=1}^n \theta_i \kappa_\sigma(\bx_i,\cdot)$. As mentioned in the introduction, this linear expansion grows indefinitely as $n$ increases; hence a sparsification strategy has to be adopted to keep the expansion's size low. In this paper, we will employ a very simple and effective strategy, which is based on the quantization of the input space \cite{QKLMS}. At each iteration. the algorithm determines whether the new point, $\bx_n$, is to be included to the list of the $M$ expansion centers, i.e., the \textit{dictionary} $C$, or not, based on its distance from $C$. If this distance is larger than a user-defined parameter $\delta$ (the \textit{quantization size}), then $\bx_n$ is inserted to $C$, otherwise the coefficient of the center that is closest to $\bx_n$ is updated. The resulting algorithm is called QKLMS:
\begin{itemize}
\item Set $f =0, C=\emptyset$, $M=0$. Select the step-size $\mu$, the parameter of the kernel $\sigma$ and the quantization size $\epsilon$.
\item for $n=1,2,\dots$ do:
\begin{enumerate}
\item Compute system's output: $\hat y_n = f(\bx_n)$.
\item Compute the error: $e_n = y_n - \hat y_n$.
\item Compute $d_k = \|\bx_n - \bc_k\|^2$, $k=1,\dots M$.
\item Find $d_{\textrm{min}} = \min\{d_k, k=1,\dots M\}$ and $k_{\textrm{min}} = \textrm{argmin}\{d_k, k=1,\dots M\}$.
\item If $d_{\min} < \epsilon$ then $\theta_{k_{\min}} = \theta_{k_{\min}} + \mu e_n$.
\item else $C = C \bigcup \{\bx_n\}$, $M=M+1$, $\theta_M = \mu e_n$.
\end{enumerate}
\end{itemize}
Note that, there are other sparsification strategies that can be applied, as it has been mentioned in the introduction. The difference is in the different criteria used to include (or not) a specific center into the dictionary. The QKLMS is among the most effective strategies and in the following it will be used as a representative of these methods. Results with other sparsification methods follow similar trends.

\section{Approximating the kernel with Random Fourier Features}\label{SEC:RFF}
The standard implementations of KLMS can be viewed as a two step procedure. Firstly, the input data, $\bx_n$, are mapped to an infinite dimensional RKHS, $\cH$, using an implicit map $\Phi(\bx_n) = \kappa(\bx_n, \cdot)$, and then the standard LMS rationale is applied to the transformed data pairs, i.e. $(\Phi(\bx_n), y_n)$, taking into account the so called \textit{kernel trick}, i.e., $\kappa(\bx_n, \bx_m) = \langle \Phi(\bx_n), \Phi(\bx_m) \rangle_\cH$, to evaluate the respective inner products. However, as it has been discussed in Section \ref{SEC:KLMS}, this leads to a solution that is expressed in terms of kernel functions, whose number keeps growing. Instead of relying on the implicit lifting provided by the kernel trick, Rahimi and Recht in \cite{RahimiRecht} proposed to map the input data to a low-dimensional Euclidean space using a randomized feature map $\bz: \R^d \rightarrow \R^D$, so that the kernel evaluations can be approximated as $\kappa(\bx_n, \bx_m) \approx \bz(\bx_n)^T \bz(\bx_m)$.

As $\bz$ is a finite dimensional lifting, direct fast linear methods can be applied to the transformed data (unlike the kernel's lifting $\Phi$, which requires special treatment). Hence, if one models the system's output as $\hat y_n = \btheta^T\bz(\bx_n)$, the standard linear LMS rationale can be applied directly to estimate the solution $\btheta\in\R^D$ at each iteration. The following theorem plays a key role in this procedure.

\begin{theorem}\label{THE:rff}
Consider a shift-invariant positive definite kernel $\kappa(\bx-\by)$ defined on $\R^d$ and its Fourier transform $p(\bomega) = \frac{1}{(2\pi)^d}\int_{\R^d} \kappa(\bdelta) e^{-i\bomega^T\bdelta} d\bdelta$, which (according to Bochner's theorem) it can be regarded as a \textbf{probability density} function. Then, defining $z_{\bomega, b}(\bx) = \sqrt{2}\cos(\bomega^T\bx + b)$, it turns out that
\begin{align}
\kappa(\bx-\by) = E_{\bomega, b}[z_{\bomega, b}(\bx) z_{\bomega, b}(\by)],
\end{align}
where $\bomega$ is drawn from $p$ and $b$ from the uniform distribution on $[0,2\pi]$.
\end{theorem}

Following Theorem \ref{THE:rff}, we choose to approximate $\kappa(\bx_n-\bx_m)$ using $D$ random Fourier features, $\bomega_1, \bomega_2, \dots, \bomega_D$, (drawn from $p$) and $D$ random numbers, $b_1, b_2, \dots, b_D$ (drawn uniformly from $[0, 2\pi]$) that define a sample average (a similar rationale as the one used in Monte Carlo Methods; for Gaussian kernels such sampling is trivial):
\begin{align}\label{EQ:approx1}
\kappa(\bx_n-\bx_m) \approx \frac{1}{D}\sum_{i=1}^D z_{\bomega_i, b_i}(\bu) z_{\bomega_i, b_i}(\bv).
\end{align}
Evidently, the larger $D$ is (up to a certain point), the better this approximation becomes. Details on the quality of this approximation can be found in \cite{RahimiRecht}.

\section{The Random Fourier Features Kernel LMS}\label{SEC:RFFKLMS}
In this Section, we briefly describe the proposed \textit{linearized} KLMS, which is based on the aforementioned Fourier approximation. The main results (regarding convergence and other related properties) are given without proofs due to lack of space.
Our starting point is to recast \eqref{EQ:approx1} in terms of Euclidean inner products. To that end, we define the map $\bz_\Omega:\R^d \rightarrow \R^D$ as follows:
\begin{align}
\bz_\Omega(\bu) = \sqrt{\frac{2}{D}}\left(\begin{matrix} \cos(\bomega_1^T\bu +b_1) \cr \vdots \cr
\cos(\bomega_D^T\bu +b_D) \end{matrix}\right),
\end{align}
where $\Omega$ is the $(d+1)\times D$ matrix defining the random fourier features of the respective kernel, i.e.,
\begin{align*}
\Omega = \left(
\begin{matrix}
\bomega_1 & \bomega_2 & ...  & \bomega_D\cr
b_1  & b_2 & ... & b_D
\end{matrix}
\right),
\end{align*}
provided that $\bomega$'s and $b$'s are drawn as mentioned above. Hence, the kernel function can be approximated as
\begin{align}
\kappa(\bx_n-\bx_m) \approx \bz_\Omega(\bx_n)^T \bz_\Omega(\bx_m).
\end{align}

Following this rationale, we propose a new variant of the KLMS, the RFFKLMS, which is actually a simple LMS on the transformed data, i.e. $\{(\bz_\Omega(\bx_n), y_n), n=1,2,\dots\}$. We model the input-output relationship as $\hat y_n = \btheta^T \bz_\Omega(\bx_n)$, for each $\bx_n$ and our goal is to evaluate $\btheta\in\R^D$ by minimizing the MSE, i.e., $J_n = E[e_n^2]$, at each time instant $n$. For the Gaussian kernel, which is employed throughout the paper, the respective Fourier transform is
\begin{align}\label{EQ:fourier_of_gaussian}
p(\bomega) = \left(\sigma/\sqrt{2\pi}\right)^D e^{-\frac{\sigma^2\|\bomega\|^2}{2}},
\end{align}
which is actually the multivariate Gaussian distribution with mean $\bzero$ and covariance matrix $\frac{1}{\sigma^2}\bI_D$.
The proposed algorithm is given next:
\begin{itemize}
\item Set $\btheta = \bZero$. Select the step-update $\mu$, the dimension of the new space, $D$ and the parameter of the kernel ($\sigma$).
\item Draw $D$ samples from $p(\bomega)$ and $D$ numbers uniformly in $[0, 2\pi]$.
\item for $n=1,2,\dots$ do:
\begin{enumerate}
\item Compute system's output: $\hat y_n = \btheta^T \bz_\Omega(\bx_n)$.
\item Compute the error: $e_n = y_n - \hat y_n$.
\item $\btheta_{n+1} = \btheta_{n} + \mu e_n \bz_\Omega(\bx_n)$.
\end{enumerate}
\end{itemize}
It is a matter of elementary algebra to conclude that after $n-1$ steps, the algorithm will give the following solution: $\btheta = \mu\sum_{k=1}^{n-1} e_k \bz_{\Omega}(\bx_k)$, which leads us to conclude that RFFKLMS will produce approximately the same system's output with the standard KLMS (provided that $D$ is sufficiently large), since
\begin{align}
\hat y_{n} = \mu\sum_{k=1}^{n-1} e_k \bz_{\Omega}(\bx_k)^T \bz_{\Omega}(\bx_{n}) \approx \mu\sum_{k=1}^{n-1} e_k \kappa_\sigma(\bx_k, \bx_{n}).
\end{align}
However, the major difference is that RFFKLMS provides a single vector $\btheta$ of fixed dimensions, instead of a growing expansion of kernel functions.

To study the convergence properties of RFFKLMS, we will assume henceforth that the data pairs are generated  by
\begin{align}
y_n = \sum_{m=1}^M a_m\kappa(\bc_m, \bx_n) + \eta_n,\label{EQ:model}
\end{align}
where $\bc_1,\dots,\bc_M$ are fixed centers, $\bx_n$ are zero-mean i.i.d, samples drawn from the Gaussian distribution with covariance matrix $\sigma_x^2\bI_d$ and $\eta_n$ are i.i.d. noise samples drawn from $\cN(0, \sigma_\eta^2)$. In this setting it is not difficult to prove that the optimal solution is given by
\begin{align}\label{EQ:optimal_solution}
\btheta_{\textrm{opt}} = \textrm{argmin}{E[e_n^2]} = Z_C \cdot \ba + R_{zz}^{-1} E[\eta'_n \cdot \bz_{\Omega}(\bx_n)],
\end{align}
where $Z_C = \left(\bz_{\Omega}(\bc_1), \dots, \bz_{\Omega}(\bc_M)\right)^T$, $\ba = (a_1, \dots, a_M)^T$, $R_{zz} = E[\bz_{\Omega}(\bx_n)\bz_{\Omega}(\bx_n)^T]$ and $\eta'_n$ is the approximation error between the noise-free component of $y_n$ (evaluated only by the linear kernel expansion of \eqref{EQ:model}) and the approximation of this component using random Fourier features, i.e., $\eta_n = \sum_{m=1}^M a_m \kappa(\bc_m,\bx_n) - \sum_{m=1}^M \bz_{\Omega}(\bc_m)^T\bz_{\Omega}(\bx_n)$. Note that this error can be made very small for sufficiently large $D$ \cite{RahimiRecht}; thus, it can be eventually dropped out. Furthermore, sufficient conditions so that $R_{zz}$ is a strictly positive definite matrix (hence invertible) have been obtained. These are summarized by:

\begin{lemma}
Consider a selection of samples $\bomega_1, \bomega_2, \dots, \bomega_D$, drawn from \eqref{EQ:fourier_of_gaussian} such that $\bomega_i\not=\bomega_j$, for any $i\not=j$. Then, the matrix $R_{zz} = E[\bz_{\Omega}(\bx_n)\bz_{\Omega}(\bx_n)^T]$ is strictly positive definite.
\end{lemma}

\noindent It is also possible, for $\bx_n\sim \cN(\bzero, \sigma_X \bI_d)$, to explicitly evaluate the entries of $R_{zz}$:
\begin{align*}
r_{i,j} =& \frac{1}{2} \exp\left(\frac{-\|\bomega_i - \bomega_j\|^2\sigma_X^2}{2}\right)\cos(b_i - b_j) \\
&+ \frac{1}{2}\exp\left(\frac{-\|\bomega_i + \bomega_j\|^2\sigma_X^2}{2}\right)\cos(b_i + b_j).
\end{align*}

As expected, the eigenvalues of $R_{zz}$ play a pivotal role in the convergence's study of the algorithm. In the case where $R_{zz}$ is a strictly positive definite matrix, its eigenvalues satisfy $0<\lambda_1\leq\lambda_2\leq\dots\leq\lambda_D$. Applying similar assumptions as in the case of the standard LMS, we can prove the following results.
\begin{proposition}\label{THE:convergence}
For datasets generated by \eqref{EQ:model} we have:
\begin{enumerate}
\item If the step update parameter satisfies $0<\mu<2/\lambda_D$, then  RFFKLSM converges in the mean, i.e., $E[\btheta_n - \btheta_{\textrm{opt}}]\rightarrow \bZero$.
\item The optimal MSE (which it is achieved when one replaces $\btheta_n$ with $\btheta_{\textrm{opt}}$) is given by $$J_n^{\textrm{opt}} = \sigma_\eta^2 + E[\eta'_n]  - E[\eta'_n\bz_{\Omega}(\bx_n)]R_{zz}^{-1}E[\eta'_n\bz_{\Omega}(\bx_n)^T].$$
For large enough $D$, we have $J_n^{\textrm{opt}} \approx \sigma_\eta^2$.
\item The excess MSE is given by $J_n^{\textrm{ex}} = J_n - J_n^{\textrm{opt}}=\trace\left(R_{zz}A_n\right)$, where $A_n = E[(\btheta_n - \btheta_{\textrm{opt}})(\btheta_n - \btheta_{\textrm{opt}})^T]$.
\item If the step update parameter satisfies $0<\mu<1/\lambda_D$, then  $A_n$ converges. For large enough $n$ and $D$ we can approximate $A_n$'s evolution as $A_{n+1} \approx A_n - \mu\left(R_{zz}A_n + A_n R_{zz}\right) + \mu^2\sigma_\eta^2 R_{zz}$. Using this model we can approximate the steady-state MSE ($\approx \trace\left(R_{zz}A_n\right) + \sigma_\eta^2$).
\end{enumerate}
\end{proposition}

\begin{figure}[t]

\begin{minipage}[b]{.48\linewidth}
  \centering
  \centerline{\epsfig{figure=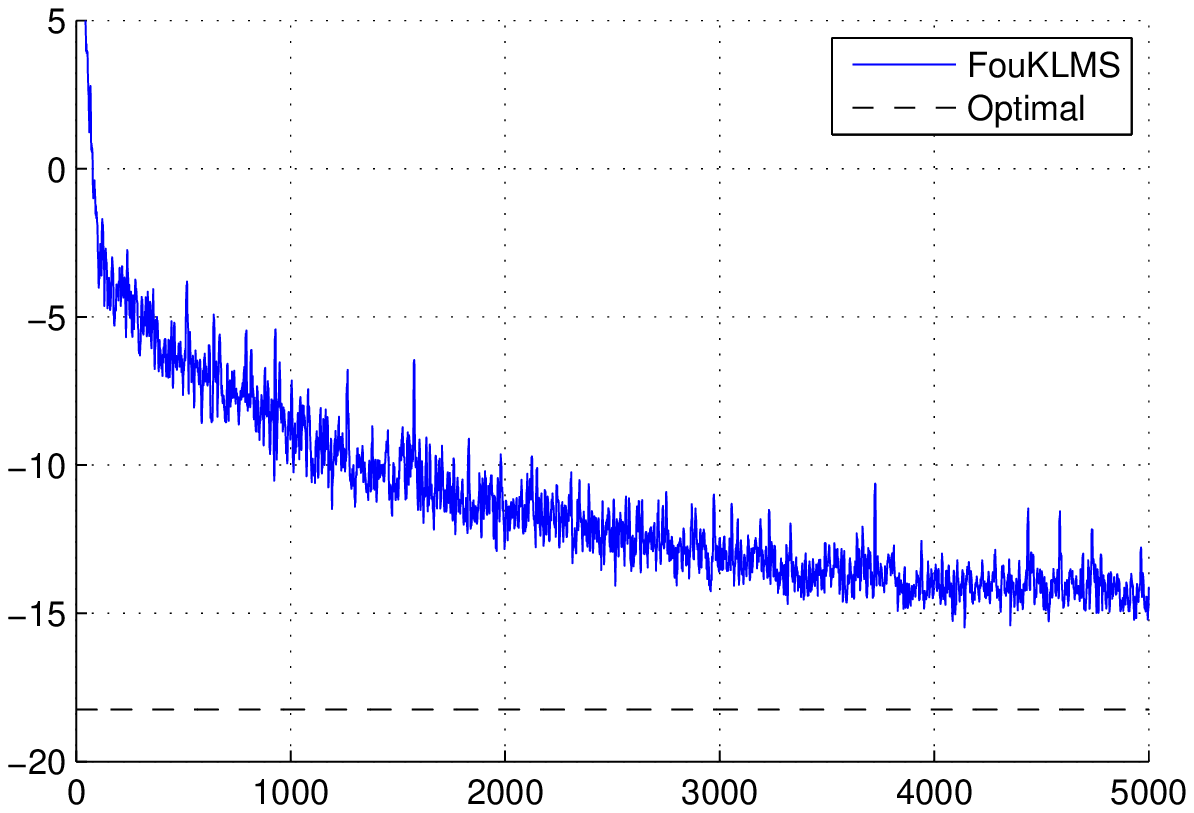,width=3.5cm}}
  \centerline{(a) $D=100$}\medskip
\end{minipage}
\hfill
\begin{minipage}[b]{0.48\linewidth}
  \centering
  \centerline{\epsfig{figure=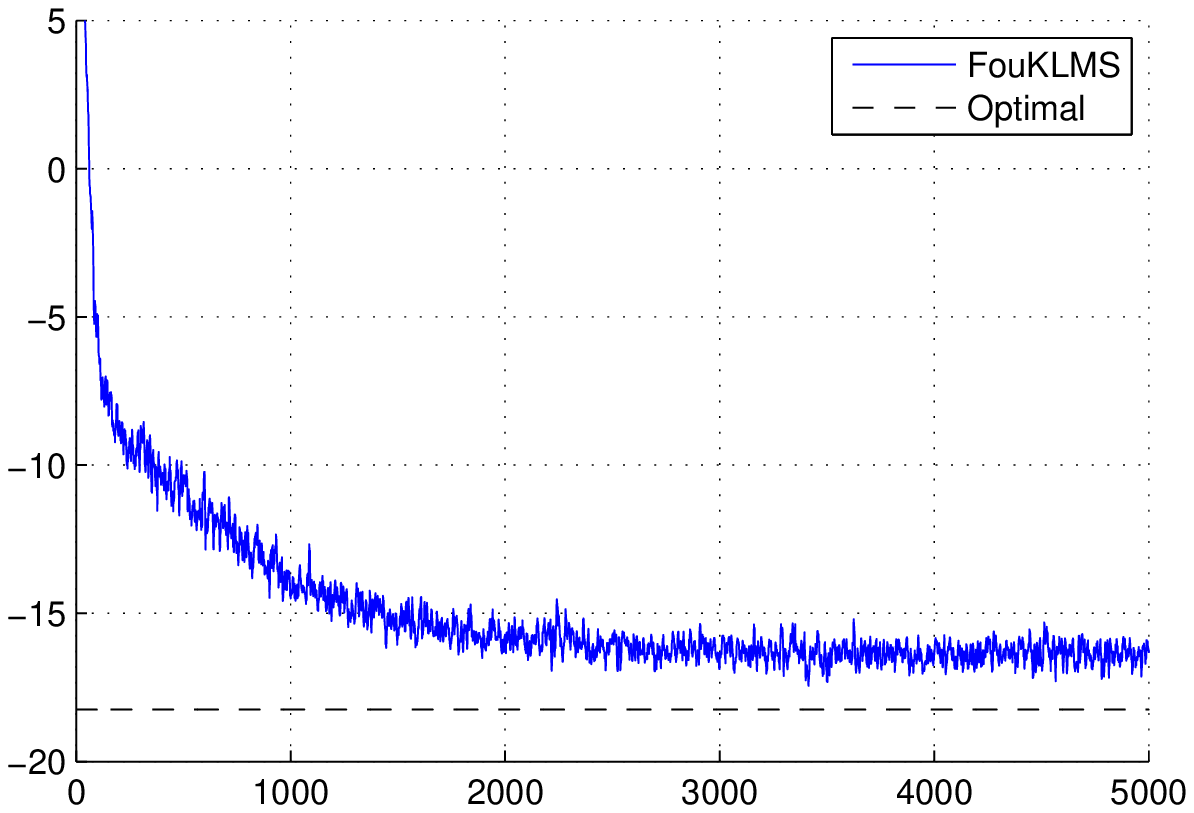,width=3.5cm}}
  \centerline{(b) $D=500$}\medskip
\end{minipage}

\begin{minipage}[b]{.48\linewidth}
  \centering
  \centerline{\epsfig{figure=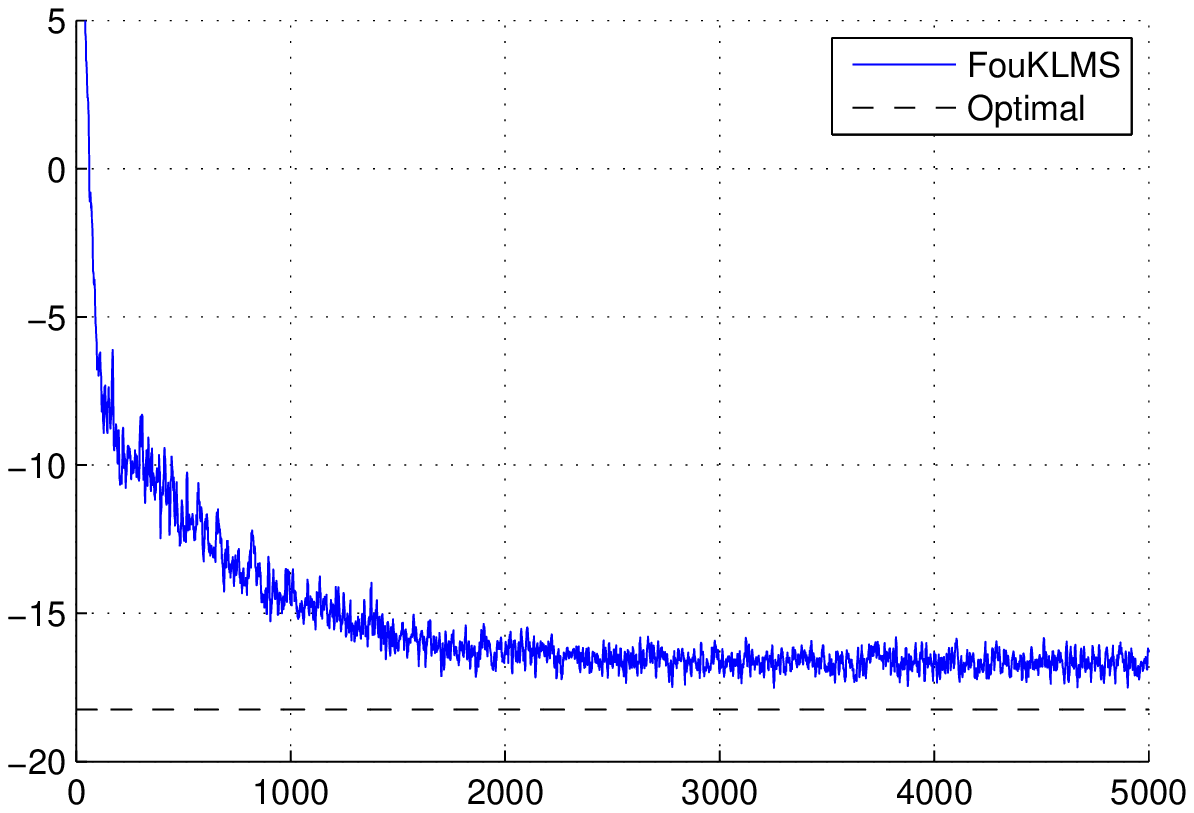,width=3.5cm}}
  \centerline{(a) $D=1000$}\medskip
\end{minipage}
\hfill
\begin{minipage}[b]{0.48\linewidth}
  \centering
  \centerline{\epsfig{figure=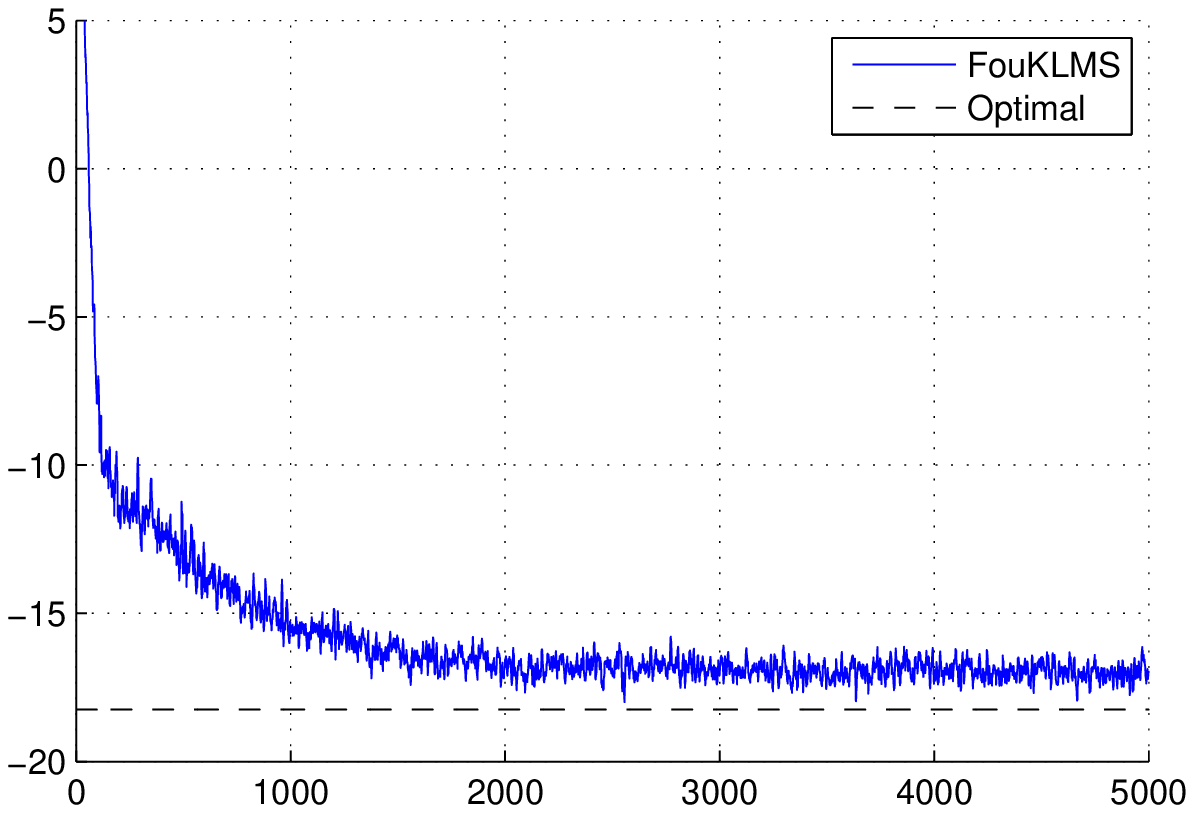,width=3.5cm}}
  \centerline{(b) $D=5000$}\medskip
\end{minipage}
\caption{Simulations of RFFKLMS (with various values of $D$) applied on data pairs generated by \eqref{EQ:model}. The results are averaged over $100$ runs. The horizontal dashed line in the figure represents the approximation of the steady-state MSE given in theorem \ref{THE:convergence}.}
\label{FIG:kernel_expansion}
\end{figure}

\section{Simulations}\label{SEC:experiments}
In this Section, we present examples to illustrate the performance of the proposed algorithm and compare its behavior to the QKLMS. In all experiments, we use the same kernel parameter, i.e., $\sigma$, for both RFFKLMS and QKLMS as well as the same step-update parameter $\mu$. The quantization parameter $\epsilon$ of the QKLMS controls the size of the dictionary. If this is too large, then the dictionary will be small and the achieved MSE at steady state will be large. Typically, however, there is a value for $\epsilon$ for which the best possible MSE (almost the same as the unsparsified version) is attained at steady state, while any smaller quantization sizes provide negligible improvements (albeit at significantly increased complexity). In all experimental set-ups, we tuned $\epsilon$ (using multiple trials) so that it takes a value close to this ``optimal'', so that to take the best possible MSE at the smallest time. On the other hand, the performance of RFFKLMS depends largely on $D$, which controls the quality of the kernel approximation. Similar to the case of QKLMS, there is a value for $D$ so that RFFKLMS attains its lowest steady-state MSE, while larger values provide negligible improvements. Table \ref{TAB:times} gives the mean training times for QKLMS and RFFKLMS on a typical core i5 machine running Matlab (both algorithms were optimized for speed). We note that the complexity of the RFFKLMS is $\mathcal{O}(Dd)$, while the complexity of QKLMS is $\mathcal{O}(Md)$. Our experiments have shown that in order to obtain similar error floors, the required complexity of RFFKLMS is lower than that of QKLMS.

\subsection{Example 1. A Linear Kernel Expansion}\label{EXP:kernel_expansion}
In this set-up we generate $5000$ data pairs using \eqref{EQ:model}. The input vectors $\bx_n$ are drawn from $\cN(\bZero, \bI)$ and the noise are i.i.d. Gaussian samples with $\sigma_\eta=0.1$. The parameters of the expansion (i.e., $a_1,\dots, a_M$) are drawn from $\cN(0, 25)$, the kernel parameter $\sigma$ is set to $5$ and the step update to $\mu=1$ (this value satisfies the requirements for convergence of Theorem \ref{THE:convergence}).
Figure \ref{FIG:kernel_expansion} shows the evolution of the MSE for 100 realizations of the experiment. The algorithm reaches steady-state around $n=2000$. The attained MSE is close to the approximation given in Theorem \ref{THE:convergence} (dashed line in the figure).

\subsection{Example 2.}\label{SEC:square}
\begin{figure}[t]
\begin{minipage}[b]{0.48\linewidth}
  \centering
  \centerline{\epsfig{figure=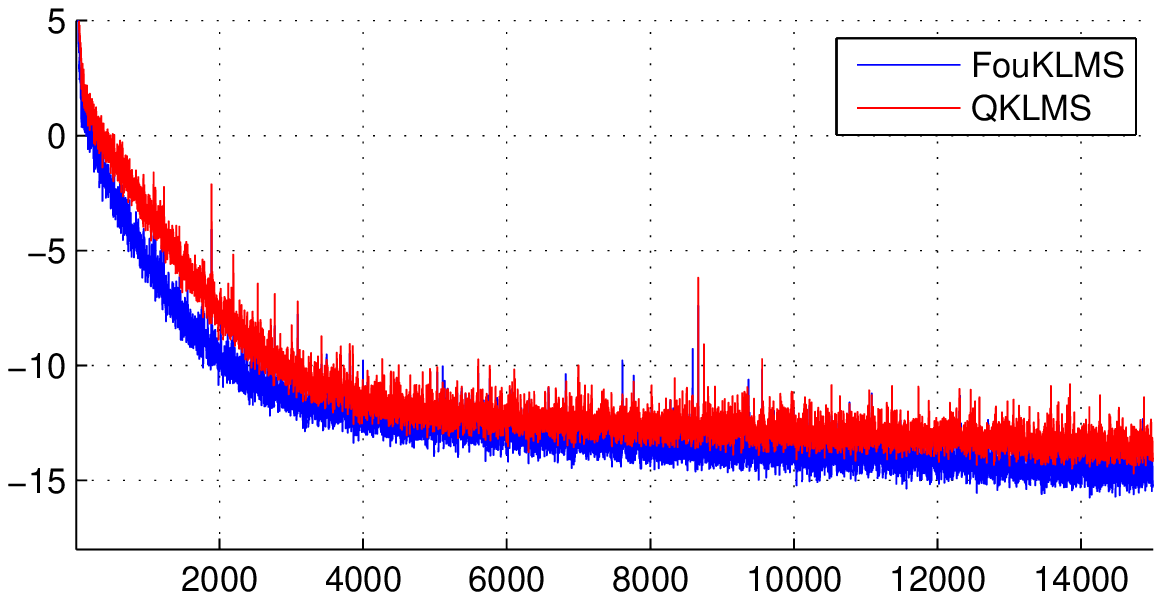,width=4cm}}
  \centerline{(a)}\medskip
\end{minipage}
\hfill
\begin{minipage}[b]{0.48\linewidth}
  \centering
  \centerline{\epsfig{figure=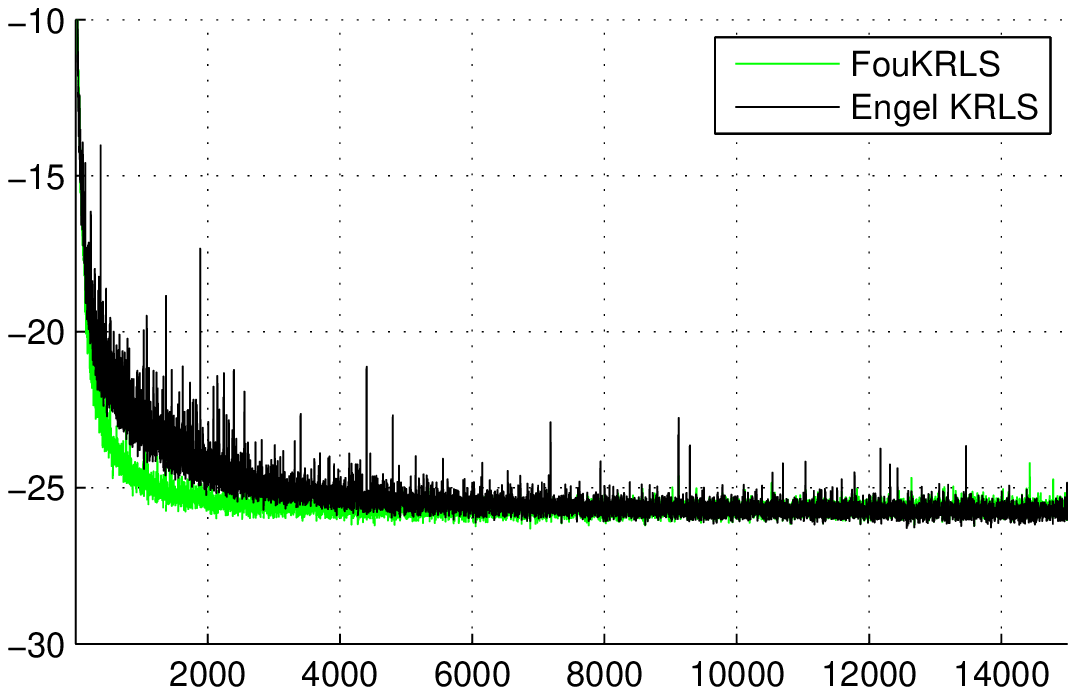,width=4cm}}
\centerline{(b)}\medskip
\end{minipage}
\caption{Monte Carlo simulations on data pairs generated as described in section \ref{SEC:square} for (a) the RFFKLMS and QKLMS, (b) the RFFKRLS and Engel's KRLS. The results are averaged over $1000$ runs.}
\label{FIG:example_square}
\vspace{-0.3cm}
\end{figure}

In this example, we adopt the following simple non-linear model:
\begin{align}
y_n = \bw_0^T\bx_n + 0.1\cdot(\bw_1^T\bx_n)^2 + \eta_n,
\end{align}
where $\eta_n$ represent zero-mean i.i.d. Gaussian noise with $\sigma_\eta=0.05$ and the coefficients of the vectors $\bw_0, \bw_1\in\R^5$ are i.i.d. samples drawn from $\cN(0,1)$. Similarly to Example 1, the kernel parameter $\sigma$ is set to $5$ and the step update to $\mu=1$. The quantization parameter of the QKLMS was set to $\epsilon=5$ (leading to an average dictionary size $M=100$) and the number of random Fourier coefficients for RFFKLMS was set to $D=300$.
Figure \ref{FIG:example_square}a shows the evolution of the MSE for both QKLMS and RFFKLMS running 1000 realizations of the experiment over $15000$ samples.

\subsection{Example 3.}\label{SEC:par1}
\begin{figure}[t]

\begin{minipage}[b]{0.48\linewidth}
  \centering
  \centerline{\epsfig{figure=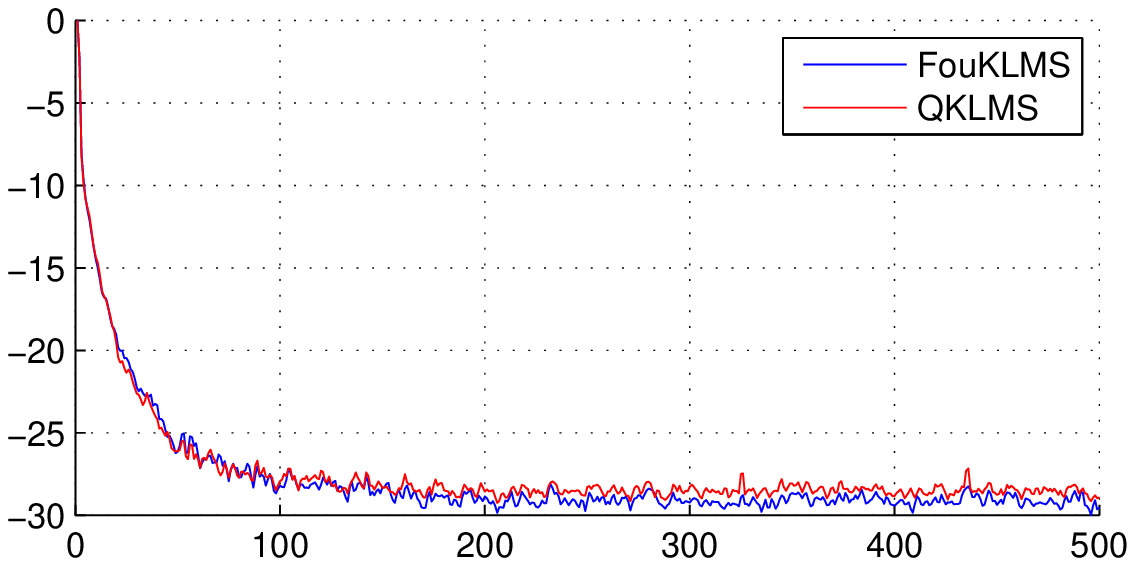,width=4cm}}
\centerline{(a)}\medskip
\end{minipage}
\hfill
\begin{minipage}[b]{0.48\linewidth}
  \centering
  \centerline{\epsfig{figure=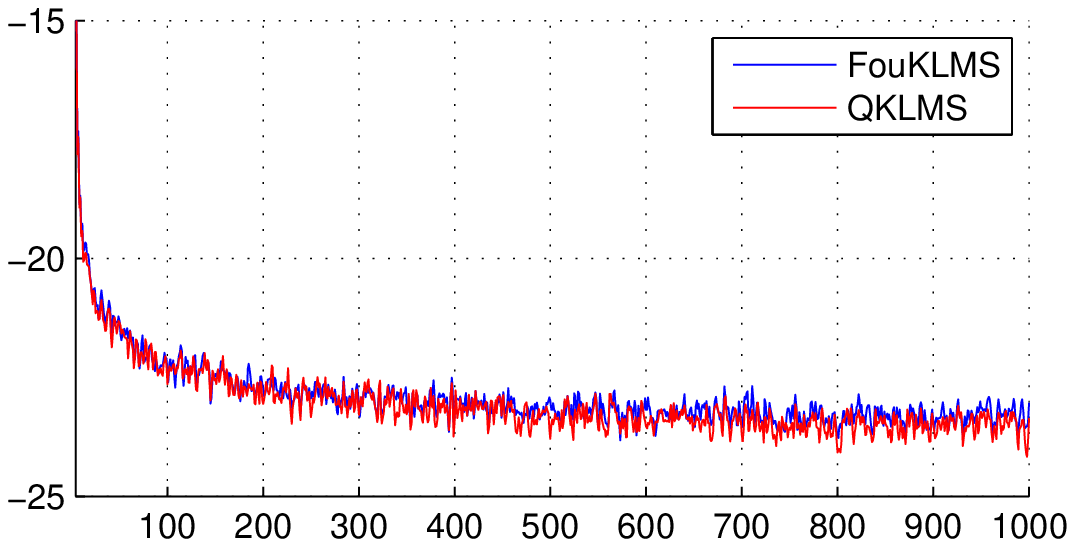,width=4cm}}
\centerline{(b)}\medskip
\end{minipage}
\caption{Monte Carlo simulation of RFFKLMS and QKLMS applied on data pairs generated as described (a) in section \ref{SEC:par1} and (b) in section \ref{SEC:par2}. The results are averaged over $1000$ runs.}
\label{FIG:example_par1}
\vspace{-0.2cm}
\end{figure}

Here we adopt the following chaotic series model \cite{Stochastic_KLMS}:
\begin{align*}
d_n = \frac{d_{n-1}}{1+d^2_{n-1}} + u^3_{n-1}, \quad
y_n = d_n + \eta_n,
\end{align*}
where $\eta_n$ is zero-mean i.i.d. Gaussian noise with $\sigma_\eta=0.01$ and $u_n$ is also zero-mean i.i.d. Gaussian with $\sigma_u = 0.15$. The kernel parameter $\sigma$ is set to $0.05$ and the step update to $\mu=1$. We have also initialized $d_1$ to $1$.
Figure \ref{FIG:example_par1}a shows the evolution of the MSE for both QKLMS and RFFKLMS running 1000 realizations of the experiment over $500$ samples. The quantization parameter $\epsilon$ was set to $\epsilon=0.01$ (leading to an average dictionary size $M=7$), while $D=100$.

\subsection{Example 4.}\label{SEC:par2}
The final example adopts another chaotic series model \cite{Stochastic_KLMS}:
\begin{align*}
d_n =& u_n + 0.5v_n - 0.2 d_{n-1} + 0.35 d_{n-2}, \\
\phi(d_n) =& \left\{\begin{matrix} \frac{d_n}{3(0.1 + 0.9 d_n^2)^{1/2}} & d_n\geq 0 \cr
 \frac{-d_n^2(1-\exp(0.7d_n))}{3} & d_n<0 \end{matrix} \right.,\quad  y_n =& \phi(d_n) + \eta_n,
\end{align*}
where $\eta_n$ is zero-mean i.i.d. Gaussian noise with $\sigma_\eta=0.001$, $v_n$ is also zero-mean i.i.d. Gaussian with $\sigma_v^2 = 0.0156$ and $u_n = 0.5v_n + \hat\eta_n$, where $\hat\eta_n$ is also i.i.d. Gaussian with $\sigma^2 = 0.0156$. The kernel parameter $\sigma$ is set to $0.05$ and the step update to $\mu=1$. We have also initialized $d_1, d_2$ to $1$.
Figure \ref{FIG:example_par1}b shows the evolution of the MSE for both QKLMS and RFFKLMS running 1000 realizations of the experiment over $1000$ samples. The parameter $\epsilon$ was set to $\epsilon=0.01$ (leading to $M=32$) and $D$ was set $D=100$.

\begin{table}[t]
\scriptsize
\begin{tabular}{|c|c|c|c|}
\hline
Experiment   &   QKLMS time  &  RFFKLMS time  &  QKLMS dictionary size\\\hline
Example 2    &   0.891 sec  &  0.226 sec      & $M=100$\\
Example 3    &   0.036 sec  &  0.006 sec      &  $M=7$\\
Example 4    &   0.057 sec  &  0.021 sec      &  $M=32$\\\hline
\end{tabular}
\caption{Mean training times for QKLMS and RFFKLMS.}\label{TAB:times}
\vspace{-0.2cm}
\end{table}

\section{The Random Fourier Features Kernel RLS}\label{SEC:krls}
Besides the implementation of the KLMS given in the previous sections, the rationale of the kernel approximation via random Fourier features (section \ref{SEC:RFF}) can also be applied to other online-algorithms such as the RLS. One only needs to choose the random samples $\bomega_i$, $b_i$ and replace the instances of $\bx_n$ in the standard RLS algorithm (see for example \cite{EREF2, Theo_ML}) with $\bz_{\Omega}(\bx_n)$. The resulting algorithm performs as well as the original KRLS provided by Engel \cite{Engel_2004_11231}, but it is almost twice as fast. Figure \ref{FIG:example_square}b compares the performances of RFFKRLS and Engle's KRLS on data samples created as in Example \ref{SEC:square}. The regularization parameter for the RFFKRLS was set to $\lambda=0.0001$, the forgetting factor to $\beta = 0.9995$, while the number of random features was set to $D=300$. The parameter for the ALD sparsification mechanism of Engel's KRLS was set to $\nu=0.0005$.

\section{Conclusions}\label{SEC:concl}
We presented an alternative rationale for the KLMS and KRLS based on the approximation of the kernel function with random Fourier Features. The proposed algorithms exhibit similar convergence performance to the standard KLMS/KRLS algorithms, albeit they require significantly lower implementation time (due to their simplicity). Furthermore, their ``linear'' characteristics pave the way for generalization to other settings (e.g., the distributed KLMS \cite{distributed_KLMS}).

\bibliographystyle{IEEEbib}
\bibliography{athensBIB}

\begin{thebibliography}{10}

\bibitem{Kivinen_2004_11230}
J.~Kivinen, A.~Smola, and R.~C. Williamson,
\newblock ``Online learning with kernels,''
\newblock {\em IEEE Transanctions on Signal Processing}, vol. 52, no. 8, pp.
  2165--2176, Aug. 2004.

\bibitem{Engel_2004_11231}
Y.~Engel, S.~Mannor, and R.~Meir,
\newblock ``The kernel recursive least-squares algorithm,''
\newblock {\em IEEE Transanctions on Signal Processing}, vol. 52, no. 8, pp.
  2275--2285, Aug. 2004.

\bibitem{Theo_ML}
Sergios Theodoridis,
\newblock {\em Machine Learning: A Bayesian and Optimization Perspective},
\newblock Academic Press, 2015.

\bibitem{EREF2}
K.~Slavakis, P.~Bouboulis, and S.~Theodoridis,
\newblock ``Online learning in reproducing kernel {H}ilbert spaces,''
\newblock in {\em Signal Processing Theory and Machine Learning}, Rama
  Chellappa and Sergios Theodoridis, Eds., Academic Press Library in Signal
  Processing, pp. 883--987. Academic Press, 2014.

\bibitem{Slavakis_2008_9257}
K.~Slavakis, S.~Theodoridis, and I.~Yamada,
\newblock ``On line kernel-based classification using adaptive projection
  algorithms,''
\newblock {\em IEEE Transactions on Signal Processing}, vol. 56, no. 7, pp.
  2781--2796, Jul. 2008.

\bibitem{Slavakis_2009_10648}
K.~Slavakis, S.~Theodoridis, and I.~Yamada,
\newblock ``Adaptive constrained {L}earning in {R}eproducing {K}ernel {H}ilbert
  spaces: the robust beamforming case,''
\newblock {\em IEEE Transactions on Signal Processing}, vol. 57, no. 12, pp.
  4744--4764, Dec. 2009.

\bibitem{Slavakis_2011_11460}
K.~Slavakis, P.~Bouboulis, and S.~Theodoridis,
\newblock ``Adaptive multiregression in reproducing kernel {H}ilbert spaces:
  the multiaccess {MIMO} channel case,''
\newblock {\em IEEE Transactions on Neural Networks and Learning Systems}, vol.
  23(2), pp. 260--276, 2012.

\bibitem{Vaerenberg_KRLS}
Vaerenbergh~S. V., L\'{a}zaro-Gredilla M., and Ignacio Santamar\'{i}a,
\newblock ``Kernel recursive least-squares tracker for time-varying
  regression,''
\newblock {\em IEEE Transanctions on Neural Networks and Learning Systems},
  vol. 23, no. 8, pp. 1313--1326, Aug. 2012.

\bibitem{Liu_2008_10645}
W.~Liu, P.~Pokharel, and J.~C. Principe,
\newblock ``The kernel {L}east-{M}ean-{S}quare algorithm,''
\newblock {\em IEEE Transanctions on Signal Processing}, vol. 56, no. 2, pp.
  543--554, Feb. 2008.

\bibitem{Bouboulis_2011_10643}
P.~Bouboulis and S.~Theodoridis,
\newblock ``Extension of {W}irtinger's calculus to {R}eproducing {K}ernel
  {H}ilbert spaces and the complex kernel {LMS},''
\newblock {\em IEEE Transactions on Signal Processing}, vol. 59, no. 3, pp.
  964--978, March 2011.

\bibitem{QKLMS}
Badong Chen, Songlin Zhao, Pingping Zhu, and J.C. Principe,
\newblock ``Quantized kernel least mean square algorithm,''
\newblock {\em IEEE Transactions on Neural Networks and Learning Systems}, vol.
  23, no. 1, pp. 22 --32, jan. 2012.

\bibitem{RichBerm}
C.~Richard, J.C.M. Bermudez, and P.~Honeine,
\newblock ``Online prediction of time series data with kernels,''
\newblock {\em IEEE Transactions on Signal Processing}, vol. 57, no. 3, pp.
  1058 --1067, march 2009.

\bibitem{Liu_2010_10644}
W.~Liu, J.~C. Principe, and S.~Haykin,
\newblock {\em {K}ernel {A}daptive {F}iltering},
\newblock Hoboken, NJ: Wiley, 2010.

\bibitem{Distributed_KLMS1}
R.~Mitra and V.~Bhatia,
\newblock ``The diffusion-klms algorithm,''
\newblock in {\em Information Technology (ICIT), 2014 International Conference
  on}, Dec 2014, pp. 256--259.

\bibitem{Distributed_KLMS2}
Wei Gao, Jie Chen, Cedric Richard, and Jianguo Huang,
\newblock ``Diffusion adaptation over networks with kernel least-mean-square,''
\newblock in {\em Computational Advances in Multi-Sensor Adaptive Processing
  (CAMSAP) 2015 International Workshop on}, 2015.

\bibitem{Distributed_KLMS3}
Chouvardas Symeon and Draief Moez,
\newblock ``A diffusion kernel {LMS} algorithm for nonlinear adaptive
  networks,''
\newblock in {\em ICASSP}, 2016.

\bibitem{RahimiRecht}
Rahimi. A. and Recht B.,
\newblock ``Random features for large scale kernel machines,''
\newblock in {\em Adv. Neural Inf. Process. Syst.} 2007, vol.~20, pp. 1177 --
  1184, Vancouver, BX, Canada.

\bibitem{DependentFourFeat}
Zhen Hu, Ming Lin, and Changshui Zhang,
\newblock ``Dependent online kernel learning with constant number of random
  fourier features,''
\newblock {\em IEEE Trans. Neural Netw. Learn. Syst.}, vol. 26, no. 10, pp.
  2464 -- 2476, October 2015.

\bibitem{Sparse_GP}
L\'{a}zaro-Gredila M., Quinonero-Candela J., Rasmussen~E. C., and
  Figueiras-Vidal~R. A.,
\newblock ``Sparse spectrum gaussian process regression,''
\newblock {\em Journal of Machine Learning Research}, vol. 11, pp. 1865--1881,
  2010.

\bibitem{Stochastic_KLMS}
W.D. Parreira, J.C.M. Bermudez, C.~Richard, and J.-Y. Tourneret,
\newblock ``Stochastic behavior analysis of the gaussian kernel
  least-mean-square algorithm,''
\newblock {\em Signal Processing, IEEE Transactions on}, vol. 60, no. 5, pp.
  2208--2222, May 2012.

\bibitem{distributed_KLMS}
Pantelis Bouboulis, Simos Chouvardas, and Sergios Theodoridis,
\newblock ``Efficient distributed online algorithms in {RKHS}: A random fourier
  feature perspective,''
\newblock {\em submitted}.

\end{thebibliography}

\end{document}